\documentclass{article}
\title{Solving the Course-timetabling Problem of Cairo University Using Max-SAT}
\author{Mohamed El Halaby}
\usepackage{comment}
\begin{document}

%

\maketitle

\begin{abstract}
Due to the good performance of current SAT (satisfiability) and Max-SAT (maximum satisfiability) solvers, many real-life optimization problems such as scheduling can be solved by encoding them into Max-SAT. In this paper we tackle the course timetabling problem of the department of mathematics, Cairo University by encoding it into Max-SAT. Generating timetables for the department by hand has proven to be cumbersome and the generated timetable almost always contains conflicts. We show how the constraints can be modelled as a Max-SAT instance.
\end{abstract}


%

\section{Introduction}
The satisfiability problem (SAT), which is the problem of deciding if there exists a truth assignment that satisfies a Boolean formula in conjunctive normal form (CNF), is a core problem in theoretical computer science because of its central position in complexity theory. Given a Boolean formula, SAT asks if there is an assignment to the variables of the formula such that it is satisfied (evaluates to true). SAT was the first problem proven to be $\mathcal{NP}$-complete by Cook \cite{cook1971complexity}. Each instance of an $\mathcal{NP}$-complete problem can be translated into an instance of SAT in polynomial time. This makes it very important to develop fast and efficient SAT solvers.

An important optimization of SAT is Max-SAT which asks for a truth assignment that satisfies the maximum number of clauses of a given CNF formula. Many theoretical and practical problems can be encoded into SAT and Max-SAT such as debugging \cite{safarpour2007improved}, circuits design and scheduling of how an observation satellite captures photos of Earth \cite{vasquez2001logic}, course timetabling \cite{asin2012curriculum,nader2004application,montero2001pso,maric2008timetabling}, software package upgrades \cite{janota2012packup}, routing \cite{xu2003sub,nam2004comparative}, reasoning \cite{sang2007dynamic} and protein structure alignment in bioinformatics \cite{pullan2007protein}.

This paper addresses the course timetabling problem of the department of mathematics, faculty of science, Cairo university. Generating a timetable is done every semester and with the growing number of students and the limited number of teaching staff, this is becoming a difficult task. An important aspect of this problem is that students of the department take different courses belonging to different curricula at the same time. This fact complicates the problem because the number of students taking an arbitrary set of courses is small and generating a conflict-free schedule becomes harder.

This paper is structured as follows. Section 2 gives basic definitions and background about SAT and Max-SAT. Section 3 describes our timetabling problem, problem encoding into partial Max-SAT and weighted Max-SAT and finally an example that illustrates the workings of the encoding. Section 4 summarizes our results and discusses future work. 

\section{Preliminaries}
A \textit{Boolean variable} $x$ can take one of two possible values 0 (false) or 1 (true). A \textit{literal} $l$ is a variable $x$ or its negation $\neg x$. A \textit{clause} is a disjunction of literals, i.e., $\bigvee_{i=1}^n l_i$. A \textit{CNF formula} is a conjunction of clauses \cite{da2010max}. Formally, a CNF formula $F$ composed of $k$ clauses, where each clause $C_i$ is composed of $m_i$ is defined as
$$F = \bigwedge_{i=1}^k C_i$$ where
$$C_i = \bigvee_{j=1}^{m_i} l_{i,j}$$
In this paper, a set of clauses $\{C_1,C_2,\dots,C_k\}$ is referred to as a Boolean formula. A truth assignment \textit{satisfies} a Boolean formula if it satisfies every clause.

Max-SAT is a generalization of SAT. Given a CNF formula $F$, the problem asks for a truth assignment that maximizes the number of satisfied clauses in $F$. For example, if $F=\{(x\vee\neg y),(\neg x\vee z),(y\vee z),(\neg z)\}$ then $I=\{x=0,y=0,z=0\}$ satisfies three clauses. In fact, $F$ is unsatisfiable and thus the maximum number of clauses that can be satisfied in $F$ is three.

There are important variations of Max-SAT such as \textit{partial} Max-SAT, \textit{weighted} Max-SAT and \textit{weighted partial} Max-SAT. In partial Max-SAT we have two sets of clauses, one is called \textit{hard} and another called \textit{soft} and we seek an assignment that satisfies all the hard clauses and maximize the number of the satisfied soft clauses. 

For example, for $F= Hard \cup Soft$, where $Hard=\{(x\vee\neg y),(\neg x\vee z)\}$ and $Soft=\{(y\vee z),(\neg z)\}$, the assignment $A=\{x=0,y=0,z=1\}$ satisfies all $Hard$ and leaves one clause in $Soft$ unsatisfied, namely $(\neg z)$. 

A weighted CNF formula is a set of clauses where each clause has an associated number. In weighted Max-SAT, we are concerned with finding a truth assignment that maximizes the sum of weights of satisfied clauses. The weighted partial Max-SAT problem is the combination between partial and weighted Max-SAT. Hard clauses have weight $\infty$.

As an example, the partial Max-SAT instance $F=\{((x,\neg y),\infty),((\neg x,z),\infty),$ $((y,z),3),((\neg z),4)\}$ has a solution $\{x=0,y=0,z=0\}$ that satisfies every hard clause and maximizes the sum of weights of satisfied soft clauses.

\section{Problem description and encoding}
The course timetabling problem deals with the following objects:
\begin{itemize}
\item Courses: Each course is associated with one teacher and a number of students. In our case, each course is scheduled twice a week, once for a lecture and another for either a section or laboratory work, but we will consider the lecture and laboratory work of the same course as different courses for technical purposes as the lecture is given by teachers and the laboratory work is supervised by teaching assistants.

\item Curricula: A curriculum is a set of courses constituting an area of specialization.

\item Rooms: Courses take place in rooms (lecture rooms or labs), where each room has an associated capacity. There are two kinds of rooms: lecture rooms and laboratories.

\item Timeslots: A timeslot is a specific day and hour during which a course is taught.
\end{itemize}
The course timetabling problem asks for an assignment of courses to rooms and timeslots in such a way that all the \textit{hard} constraints are met and satisfy as much \textit{soft} constraints as possible.

Because of the large number of students taking courses belonging to different curricula, a generated timetable is prone to have conflicts (with two or more courses taken by a student scheduled at the same time), which is a very important problem in the timetabling problem we are tackling. Other formulations do not take this conflict into account.

Our timetabling approach assumes that the timetable is a per-week schedule of courses. A week consists of several work days, and each work day is divided into equal-length timeslots and courses are scheduled into these timeslots.

Many of our timetabling requirements are common for many teaching institutions, however, some may vary. This allows us to use existing encodings and timetabling tools.

\subsection{Partial Max-SAT encoding}
Here, the problem is modelled as a PMax-SAT instance. We first define the following variables upon which the encoding is based, then we mention the constraints and requirements of the timetable. The notation used is the same one used in \cite{asin2012curriculum}.
\begin{enumerate}
\item $ct_{c,t}$: course $c$ takes place in timeslot $t$.
\item $cd_{c,d}$: course $c$ takes place in day $d$.
\item $cr_{c,r}$: course $c$ takes place in room $r$.
\item $kt_{k,t}$: curriculum $k$ takes place in timeslot $t$, which implies that one of the courses belonging to $k$ takes place in $t$.
\end{enumerate}
The truth values of these basic variables will determine the whole timetable. First, we show how to encode \textit{correctness requirements}, which must all be satisfied (i.e., they are hard constraints) for the generated timetable to be accepted. So, in order for the encoding to be correct, the relation between the variables must be expressed correctly.\\
\textbf{Relationship between $ct$ and $cd$}:
\begin{enumerate}
\item If some course $c$ takes place in timeslot $t$, it also takes place in the day corresponding to $t$. So, for each course c and timeslot h, the following hard clause is needed:
$$ct_{c,t} \Rightarrow cd_{c,day(t)}$$ which is equivalent to\footnote{$p \Rightarrow q$ is equivalent to $\neg p \lor q$. From this point forward we will directly write the encoding clauses using only ORs $\lor$ and ANDs $\wedge$.}
$$\neg ct_{c,t} \lor cd_{c,day(t)}$$

\item If some course $c$ takes place in a day $d$, it must also occur in some of the timeslots of $d$. So, for each course $c$ and day $d$ consisting of timeslots $t_1,t_2,\dots,t_n$, the following hard clause is needed:
$$\neg cd_{c,d} \lor ct_{c,t_1} \lor \dots \lor ct_{c,t_n}$$
\end{enumerate}
\textbf{Relationship between $ct$ and $kt$}:
\begin{itemize}
\item If a course $c$ takes place in timeslot $t$, then all the curricula $k_1,k_2,\dots,k_n$, to which $c$ belongs occur in $t$. So, for each course $c$, timeslot $t$ and curricula $k_1,k_2,\dots,k_n$ that include $c$, the following hard clauses are needed:
$$\neg ct_{c,t} \lor kt_{k_1,t}$$
$$\vdots$$
$$\neg ct_{c,t} \lor kt_{k_n,t}$$

\item If a curriculum $k$ takes place in timeslot $t$, then at least one of the courses belonging to $k$ takes place in $t$. So, for each timeslot $t$ and curriculum $k$ consisting of courses $c_1,c_2,\dots,c_n$ we need the following hard clause
$$\neg kt_{k,t} \lor (ct_{c_1,t} \lor \dots \lor ct_{c_n,t})$$
\end{itemize}

Now after the relationships between the variables have been encoded, we encode the constraints of the timetabling problem. Here we omit the details of the cardinality constraints by denoting them by \textit{exactly}, \textit{at most} and \textit{at least}. Several attempts have been made to encode the cardinality constraints into CNF such as \cite{abio2013parametric}, \cite{marques2007towards} and \cite{bailleux2003efficient}.
\begin{itemize}
\item Curriculum clashes: Courses belonging to the same curriculum must be scheduled at different timeslots. Thus, for every two distinct courses $c_1$ and $c_2$ belonging to the same curriculum, and for every timeslot $t$, the following hard clause is needed
$$\neg ct_{c_1,t} \lor \neg ct_{c_2,t}$$

\item Student registration clashes: Students are allowed to enrol themselves in any of the available courses except if the student had not passed its prerequisite. Hence, there are many cases where students register two or more courses from different curricula, e.g., a course from first year curriculum and another from the second year curriculum. The main reason for this is that a student did not pass a first year course and he or she is taking it again. So, for every timeslot $t$ and two courses $c_1$ and $c_2$ registered by a student, the following soft clause is needed
$$\neg ct_{c_1,t} \lor \neg ct_{c_2,t}$$
The reason for making the clause soft is that insisting on satisfying all these clauses is not realistic because of the limited number of teaching staff and the large number of students. 

Before the beginning of each semester, the timetable coordinators receive from students the courses they hope to register from different curricula. Not all such courses can be scheduled with no conflicts with other courses.

\item Teacher clashes: Courses taught by the same teacher must not be scheduled at the same time. So, for every timeslot $t$ and two distinct courses $c_1$ and $c_2$ taught by the same teacher, the following hard clause is needed
$$\neg ct_{c_1,t} \lor \neg ct_{c_2,t}$$
\item Room clashes: Every room can have at most one course scheduled to it at any timeslot. So, for each room $r$, timeslot $t$, and pair of distinct courses $c_1$ and $c_2$, we have
$$\neg ct_{c_1,t} \lor \neg ct_{c_2,t} \lor \neg cr_{c_1,r} \lor \neg cr_{c_2,r}$$ 

\item Timeslot unavailability: Each course can have a set of forbidden timeslots to be scheduled in. For each course $c$ with forbidden timeslots $t_1,t_2,\dots,t_n$, the following soft clauses are needed
$$\neg ct_{c,t_1}$$
$$\vdots$$
$$\neg ct_{c,t_{n}}$$
In our case, timeslot unavailability are times during which teachers are not available to teach their courses. It is practically difficult to satisfy every such constraint and that is why they are declared soft. 

\item Room capacity: Every course is preferred to be scheduled into a room that fits. So, for every course $c$ having $n_c$ students and for every room with capacity $c_r$ such that $n_s > c_r$, we need the following soft clause
$$\neg cr_{c,r}$$

\item Every course must be scheduled to exactly one room. Courses with weekly laboratory work must be scheduled to a room that is a laboratory. We will denote regular lecture rooms $r_1,\dots,r_m$ and laboratory rooms $lab_1,\dots,lab_n$.
	\begin{enumerate}
	\item For each course $c$ with weekly lecture and laboratory work, we need the following hard clause for the lecture:
	$$exactly(1,\{cr_{c,r_1},\dots,cr_{c,r_m},cr_{c,lab_1},\dots,cr_{c,lab_n}\})$$
	and the following hard clause for the laboratory work: $$exactly(1,\{cr_{c,lab_1},\dots,cr_{c,lab_n}\})$$
	\item For each course $c$ with a weekly lecture and a section and no laboratory work, we need the following hard clause:
	$$exactly(2,\{cr_{c,r_1},\dots,cr_{c,r_m},cr_{c,lab_1},\dots,cr_{c,lab_n}\})$$
	\end{enumerate}
	The clauses in the previous two cases are SAT encodings of the cardinality constraint \textit{exactly}, which requires that exactly $k$ out of $l$ literals be true.
\item Number of lectures: Each course $c$ must be scheduled twice a week; once for a lecture and another for either a section (e.g., for solving exercises) or laboratory work (e.g., for programming exercises). So, for every course $c$ we need the following hard clause
$$exactly(2,\{ct_{c,t_1},\dots,ct_{c,t_n}\})$$ 
\end{itemize}

\subsection{Weighted partial Max-SAT encoding}
Not all the soft constraints are of the same importance and a practical solution must take this fact into account. The previous PMax-SAT encoding can be extended to associate the encoding clauses of an important constraint with a big weight and the less important constraints with smaller weights. For every soft constraint a positive weight is going to be assigned:
\begin{enumerate}
\item Student registration clashes: Courses that belong to different curricula and desired by more students will receive larger weight. So, for every two courses $c_1$ and $c_2$ needed by a number of students $n$ belonging to two different curricula, and for every timeslot $t$, the following soft clause is needed with weight $n$
$$(\neg ct_{c_1,t} \lor ct_{c_2,t})$$

\item Timeslot unavailability: Each clause representing a timeslot unavailability will receive a weight of 10. This constraint is considered less important than student registration clashes.

\item Room capacity. Each violation of this constraint has a cost of 1 per each student that does not fit. For example if the capacity of a room $r$ is 50 and the number of students in a course $c$ is 60, then the weight assigned to $(\neg cr_{c,r})$ is 10.
\end{enumerate}

\subsection{Example}
The following is a simple example to demonstrate the partial Max-SAT encoding.
\begin{itemize}
\item Days: $\{d_1,d_2,d_3\}$.

\item Timeslots: $\{t_1,t_2,t_3,t_4,t_5\}$, where $day(t_1)=d_1$, $day(t_2)=d_1$, $day(t_3)=d_2$, $day(t_4)=d_2$ and $day(t_5)=d_3$.

\item Courses: $\{CS101, CS202, M271, CS304, CS305, CS402,\\ CS408\}$. 
\begin{center}
\tiny{
\begin{tabular}{|c|c|c|c|c|}
\hline \textbf{Course ID} & \textbf{Name} & \textbf{Curriculum} & \textbf{Teacher} & \textbf{Forbidden timeslots} \\ 
\hline CS101 & Introduction to Computers &$k_1$ & $I_1$ & $\{t_1,t_2\}$ \\ 
\hline CS202 & Data Structures and Algorithms & $k_2$ & Rasha & $\{t_5\}$ \\ 
\hline M271 & Newtonian Mechanics  & $k_2$ & Enaam & $\emptyset$ \\ 
\hline CS305 & Algorithms & $k_3$ & Hassan & $\{t_1\}$ \\ 
\hline CS304 & Computer Graphics & $k_3$ & Hassan & $\{t_1\}$ \\ 
\hline CS402 & Cryptography & $k_4$ & Rasha & $\{t_4,t_5\}$ \\ 
\hline CS408 & Artificial Intelligence & $k_4$ & Alaa & $\emptyset$ \\ 
\hline 
\end{tabular}
}
\end{center}

\item Curricula: $\{k_1=\{CS101\}, k_2=\{CS202, M271\}, k_3=\{CS304, CS305\}, k_4=\{CS402, CS408\}\}$.

\item Student registration information:
\small{
\begin{center}
\begin{tabular}{|c|c|}
\hline \textbf{Courses desired} & \textbf{Number of students} \\ 
\hline $\{CS101\}$ & 55 \\
\hline $\{CS101,M271\}$ & 20 \\ 
\hline $\{M271,CS202\}$ & 50 \\
\hline $\{CS304,CS305\}$ & 64 \\
\hline $\{CS305, M271\}$ & 15 \\
\hline $\{CS402,CS408\}$ & 50 \\
\hline $\{CS304,CS402\}$ & 10 \\
\hline $\{CS408,M271\}$ & 5 \\
\hline
\end{tabular}

\begin{tabular}{|c|c|}
\hline \textbf{Course} & \textbf{Number of students}\\
\hline CS101 & 20 \\
\hline CS202 & 50 \\
\hline M271 & 90 \\
\hline CS304 & 74 \\
\hline CS305 & 79 \\
\hline CS402 & 60 \\
\hline CS402 & 55 \\
\hline 
\end{tabular}
\end{center}
}

\item Rooms: $\{r_1,r_2,lab1,lab2\}$. 
\begin{center}
\begin{tabular}{|c|c|}
\hline \textbf{Rooms} & \textbf{Capacity} \\ 
\hline $r_1$ & 50 \\ 
\hline $r_2$ & 100 \\ 
\hline lab1 & 50 \\ 
\hline lab2 & 100 \\ 
\hline 
\end{tabular} 
\end{center}
\end{itemize}

\subsubsection{Clauses}
\begin{enumerate}
\item Relationship between $ct$ and $cd$.
The clauses are grouped by course.\\
CS101:
\small{
$$(\neg ct_{CS101, t_1} \lor cd_{CS101, d_1}),(\neg ct_{CS101, t_2} \lor cd_{CS101, d_1})$$
$$(\neg ct_{CS101, t_3} \lor cd_{CS101, d_2}),(\neg ct_{CS101, t_4} \lor cd_{CS101, d_2})$$
$$(\neg ct_{CS101, t_5} \lor cd_{CS101, d_3}),(\neg cd_{CS101,d_1} \lor ct_{CS101,t_1} \lor ct_{CS101,t_2})$$
$$(\neg cd_{CS101,d_2} \lor ct_{CS101,t_3} \lor ct_{CS101,t_4}),(\neg cd_{CS101,d_3} \lor ct_{CS101,t_5})$$
}
CS202:
$$(\neg ct_{CS202, t_1} \lor cd_{CS202, d_1}),(\neg ct_{CS202, t_2} \lor cd_{CS202, d_1})$$
$$(\neg ct_{CS202, t_3} \lor cd_{CS202, d_2}),(\neg ct_{CS202, t_4} \lor cd_{CS202, d_2})$$
$$(\neg ct_{CS202, t_5} \lor cd_{CS202, d_3}),(\neg cd_{CS202,d_1} \lor ct_{CS202,t_1} \lor ct_{CS202,t_2})$$
$$(\neg cd_{CS202,d_2} \lor ct_{CS202,t_3} \lor ct_{CS202,t_4}),(\neg cd_{CS202,d_3} \lor ct_{CS202,t_5})$$
M271:
$$(\neg ct_{M271, t_1} \lor cd_{M271, d_1}),(\neg ct_{M271, t_2} \lor cd_{M271, d_1})$$
$$(\neg ct_{M271, t_3} \lor cd_{M271, d_2}),(\neg ct_{M271, t_4} \lor cd_{M271, d_2})$$
$$(\neg ct_{M271, t_5} \lor cd_{M271, d_3}),(\neg cd_{M271,d_1} \lor ct_{M271,t_1} \lor ct_{M271,t_2})$$
$$(\neg cd_{M271,d_2} \lor ct_{M271,t_3} \lor ct_{M271,t_4}),(\neg cd_{M271,d_3} \lor ct_{M271,t_5})$$
CS304:
$$(\neg ct_{CS304, t_1} \lor cd_{CS304, d_1}),(\neg ct_{CS304, t_2} \lor cd_{CS304, d_1})$$
$$(\neg ct_{CS304, t_3} \lor cd_{CS304, d_2}),(\neg ct_{CS304, t_4} \lor cd_{CS304, d_2})$$
$$(\neg ct_{CS304, t_5} \lor cd_{CS304, d_3}),(\neg cd_{CS304,d_1} \lor ct_{CS304,t_1} \lor ct_{CS304,t_2})$$
$$(\neg cd_{CS304,d_2} \lor ct_{CS304,t_3} \lor ct_{CS304,t_4}),(\neg cd_{CS304,d_3} \lor ct_{CS304,t_5})$$
CS305:
$$(\neg ct_{CS305, t_1} \lor cd_{CS305, d_1}),(\neg ct_{CS305, t_2} \lor cd_{CS305, d_1})$$
$$(\neg ct_{CS305, t_3} \lor cd_{CS305, d_2}),(\neg ct_{CS305, t_4} \lor cd_{CS305, d_2})$$
$$(\neg ct_{CS305, t_5} \lor cd_{CS305, d_3}),(\neg cd_{CS305,d_1} \lor ct_{CS305,t_1} \lor ct_{CS305,t_2})$$
$$(\neg cd_{CS305,d_2} \lor ct_{CS305,t_3} \lor ct_{CS305,t_4}),(\neg cd_{CS305,d_3} \lor ct_{CS305,t_5})$$
CS402:
$$(\neg ct_{CS402, t_1} \lor cd_{CS402, d_1}),(\neg ct_{CS402, t_2} \lor cd_{CS402, d_1})$$
$$(\neg ct_{CS402, t_3} \lor cd_{CS402, d_2}),(\neg ct_{CS402, t_4} \lor cd_{CS402, d_2})$$
$$(\neg ct_{CS402, t_5} \lor cd_{CS402, d_3}),(\neg cd_{CS402,d_1} \lor ct_{CS402,t_1} \lor ct_{CS402,t_2})$$
$$(\neg cd_{CS402,d_2} \lor ct_{CS402,t_3} \lor ct_{CS402,t_4}),(\neg cd_{CS402,d_3} \lor ct_{CS402,t_5})$$
CS408:
$$(\neg ct_{CS408, t_1} \lor cd_{CS408, d_1}),(\neg ct_{CS408, t_2} \lor cd_{CS408, d_1})$$
$$(\neg ct_{CS408, t_3} \lor cd_{CS408, d_2}),(\neg ct_{CS408, t_4} \lor cd_{CS408, d_2})$$
$$(\neg ct_{CS408, t_5} \lor cd_{CS408, d_3}),(\neg cd_{CS408,d_1} \lor ct_{CS408,t_1} \lor ct_{CS408,t_2})$$
$$(\neg cd_{CS408,d_2} \lor ct_{CS408,t_3} \lor ct_{CS408,t_4}),(\neg cd_{CS408,d_3} \lor ct_{CS408,t_5})$$

\item Relationship between $ct$ and $kt$. The clauses are grouped by course:\\
CS101:\\
$(\neg ct_{CS101,t_1} \lor kt_{k_1,t_1}),(\neg ct_{CS101,t_2} \lor kt_{k_1,t_2}),(\neg ct_{CS101,t_3} \lor kt_{k_1,t_3}), (\neg ct_{CS101,t_4} \lor kt_{k_1,t_4}),(\neg ct_{CS101,t_5} \lor kt_{k_1,t_5})$\\

CS202:\\
$(\neg ct_{CS202,t_1} \lor kt_{k_2,t_1}),(\neg ct_{CS202,t_2} \lor kt_{k_2,t_2}),(\neg ct_{CS202,t_3} \lor kt_{k_2,t_3}), (\neg ct_{CS202,t_4} \lor kt_{k_2,t_4}),(\neg ct_{CS202,t_5} \lor kt_{k_2,t_5})$\\

M271:\\
$(\neg ct_{M271,t_1} \lor kt_{k_2,t_1}),(\neg ct_{M271,t_2} \lor kt_{k_2,t_2}),(\neg ct_{M271,t_3} \lor kt_{k_2,t_3}), (\neg ct_{M271,t_4} \lor kt_{k_2,t_4}),(\neg ct_{M271,t_5} \lor kt_{k_2,t_5})$\\

CS304:\\
$(\neg ct_{CS304,t_1} \lor kt_{k_3,t_1}),(\neg ct_{CS304,t_2} \lor kt_{k_3,t_2}),(\neg ct_{CS304,t_3} \lor kt_{k_3,t_3}), (\neg ct_{CS304,t_4} \lor kt_{k_3,t_4}),(\neg ct_{CS304,t_5} \lor kt_{k_3,t_5})$\\

CS305:\\
$(\neg ct_{CS305,t_1} \lor kt_{k_3,t_1}),(\neg ct_{CS305,t_2} \lor kt_{k_3,t_2}),(\neg ct_{CS305,t_3} \lor kt_{k_3,t_3}), (\neg ct_{CS305,t_4} \lor kt_{k_3,t_4}),(\neg ct_{CS305,t_5} \lor kt_{k_3,t_5})$\\

CS402:\\
$(\neg ct_{CS402,t_1} \lor kt_{k_4,t_1}),(\neg ct_{CS402,t_2} \lor kt_{k_4,t_2}),(\neg ct_{CS402,t_3} \lor kt_{k_4,t_3}), (\neg ct_{CS402,t_4} \lor kt_{k_4,t_4}),(\neg ct_{CS402,t_5} \lor kt_{k_4,t_5})$\\

CS408:\\
$(\neg ct_{CS408,t_1} \lor kt_{k_4,t_1}),(\neg ct_{CS408,t_2} \lor kt_{k_4,t_2}),(\neg ct_{CS408,t_3} \lor kt_{k_4,t_3}), (\neg ct_{CS408,t_4} \lor kt_{k_4,t_4}),(\neg ct_{CS408,t_5} \lor kt_{k_4,t_5})$\\

\item Curriculum clashes. The clauses are grouped courses of the same curriculum:  \\
CS202 and M271:\\
$(\neg ct_{CS202,t_1} \lor \neg ct_{M271,t_1}),(\neg ct_{CS202,t_2} \lor \neg ct_{M271,t_2}),(\neg ct_{CS202,t_3} \lor \neg ct_{M271,t_3}), (\neg ct_{CS202,t_4} \lor \neg ct_{M271,t_4}),(\neg ct_{CS202,t_5} \lor \neg ct_{M271,t_5})$\\

CS304 and CS305:\\
$(\neg ct_{CS304,t_1} \lor \neg ct_{CS305,t_1}),(\neg ct_{CS304,t_2} \lor \neg ct_{CS305,t_2}),(\neg ct_{CS304,t_3} \lor \neg ct_{CS305,t_3}), (\neg ct_{CS304,t_4} \lor \neg ct_{CS305,t_4}),(\neg ct_{CS304,t_5} \lor \neg ct_{CS305,t_5})$\\

CS402 and CS408:\\
$(\neg ct_{CS402,t_1} \lor \neg ct_{CS408,t_1}),(\neg ct_{CS402,t_2} \lor \neg ct_{CS408,t_2}),(\neg ct_{CS402,t_3} \lor \neg ct_{CS408,t_3}), (\neg ct_{CS402,t_4} \lor \neg ct_{CS408,t_4}),(\neg ct_{CS402,t_5} \lor \neg ct_{CS408,t_5})$\\

\item Student registration clashes. The clauses are grouped by each two courses that can not be scheduled at the same time:\\
CS101 and M271:
$(\neg ct_{CS101,t_1} \lor \neg ct_{M271,t_1}),(\neg ct_{CS101,t_2} \lor \neg ct_{M271,t_2}),(\neg ct_{CS101,t_3} \lor \neg ct_{M271,t_3}), (\neg ct_{CS101,t_4} \lor \neg ct_{M271,t_4}),(\neg ct_{CS101,t_5} \lor \neg ct_{M271,t_5})$\\

M271 and CS305:\\
$(\neg ct_{CS305,t_1} \lor \neg ct_{M271,t_1}),(\neg ct_{CS305,t_2} \lor \neg ct_{M271,t_2}),(\neg ct_{CS305,t_3} \lor \neg ct_{M271,t_3}), (\neg ct_{CS305,t_4} \lor \neg ct_{M271,t_4}),(\neg ct_{CS305,t_5} \lor \neg ct_{M271,t_5})$\\

M271 and CS408:\\
$(\neg ct_{CS408,t_1} \lor \neg ct_{M271,t_1}),(\neg ct_{CS408,t_2} \lor \neg ct_{M271,t_2}),(\neg ct_{CS408,t_3} \lor \neg ct_{M271,t_3}), (\neg ct_{CS408,t_4} \lor \neg ct_{M271,t_4}),(\neg ct_{CS408,t_5} \lor \neg ct_{M271,t_5})$\\

CS402 and CS304:\\
$(\neg ct_{CS304,t_1} \lor \neg ct_{CS402,t_1}),(\neg ct_{CS304,t_2} \lor \neg ct_{CS402,t_2}),(\neg ct_{CS304,t_3} \lor \neg ct_{CS402,t_3}), (\neg ct_{CS304,t_4} \lor \neg ct_{CS402,t_4}),(\neg ct_{CS304,t_5} \lor \neg ct_{CS402,t_5})$\\

\item Teacher clashes. The clauses are grouped by each pair of courses taught by the same teacher:\\
CS202 and CS402:\\
$(\neg ct_{CS202,t_1} \lor \neg ct_{CS402,t_1}),(\neg ct_{CS202,t_2} \lor \neg ct_{CS402,t_2}),(\neg ct_{CS202,t_3} \lor \neg ct_{CS402,t_3}), (\neg ct_{CS202,t_4} \lor \neg ct_{CS402,t_4}),(\neg ct_{CS202,t_5} \lor \neg ct_{CS402,t_5})$

\item Room clashes. We will not write all the clauses since they are too many. Clauses for one room, six pairs of courses and one timeslot:\\
$r_1,t_1$:
$$(\neg ct_{CS101,t_1} \lor \neg ct_{CS202,t_1} \lor \neg cr_{CS101,r_1} \lor \neg cr_{CS202,r_1}),$$
$$(\neg ct_{CS101,t_1} \lor \neg ct_{M271,t_1} \lor \neg cr_{CS101,r_1} \lor \neg cr_{M271,r_1}),$$
$$(\neg ct_{CS101,t_1} \lor \neg ct_{CS304,t_1} \lor \neg cr_{CS101,r_1} \lor \neg cr_{CS304,r_1}),$$
$$(\neg ct_{CS101,t_1} \lor \neg ct_{CS305,t_1} \lor \neg cr_{CS101,r_1} \lor \neg cr_{CS305,r_1}),$$
$$(\neg ct_{CS101,t_1} \lor \neg ct_{CS402,t_1} \lor \neg cr_{CS101,r_1} \lor \neg cr_{CS402,r_1}),$$
$$(\neg ct_{CS101,t_1} \lor \neg ct_{CS408,t_1} \lor \neg cr_{CS101,r_1} \lor \neg cr_{CS408,r_1})$$

\item Timeslot unavailability. The clauses are grouped by each teacher:\\
$I_1$:\\
$$(\neg ct_{CS101,t_1}),(\neg ct_{CS101,t_2})$$
$I_2$:\\
$$(\neg ct_{CS202,t_4}),(\neg ct_{CS402,t_4}),(\neg ct_{CS402,t_5})$$
Hassan:\\
$$(\neg ct_{CS304,t_1}),(\neg ct_{CS305,t_1})$$

\item Room capacity. The clauses are grouped by courses:\\
CS101: $n_s=75$
$$(\neg cr_{CS101,r_1}),(\neg cr_{CS101,lab1})$$
M271: $n_s=90$
$$(\neg cr_{M271,r_1}),(\neg cr_{M271,lab1})$$
CS304: $n_s=74$
$$(\neg cr_{CS304,r_1}),(\neg cr_{CS304,lab1})$$
CS305: $n_s=79$
$$(\neg cr_{CS305,r_1}),(\neg cr_{CS305,lab1})$$
CS402: $n_s=60$
$$(\neg cr_{CS402,r_1}),(\neg cr_{CS402,lab1})$$
CS408: $n_s=55$
$$(\neg cr_{CS408,r_1}),(\neg cr_{CS408,lab1})$$

\item Every course must be scheduled to exactly one room. The clauses are grouped by courses:\\
CS101:
$$exactly(1,\{cr_{CS101,r_1},cr_{CS101,r_2},cr_{CS101,lab1},cr_{CS101,lab2}\})$$
$$exactly(1,\{cr_{CS101,lab1},cr_{CS101,lab2}\})$$
CS202:
$$exactly(1,\{cr_{CS202,r_1},cr_{CS202,r_2},cr_{CS202,lab1},cr_{CS202,lab2}\})$$
$$exactly(1,\{cr_{CS202,lab1},cr_{CS202,lab2}\})$$
M271:
$$exactly(2,\{cr_{M271,r_1},cr_{M271,r_2},cr_{M271,lab1},cr_{M271,lab2}\})$$
CS304:
$$exactly(1,\{cr_{CS304,r_1},cr_{CS304,r_2},cr_{CS304,lab1},cr_{CS304,lab2}\})$$
$$exactly(1,\{cr_{CS304,lab1},cr_{CS304,lab2}\})$$
CS305:
$$exactly(1,\{cr_{CS305,r_1},cr_{CS305,r_2},cr_{CS305,lab1},cr_{CS305,lab2}\})$$
$$exactly(1,\{cr_{CS305,lab1},cr_{CS305,lab2}\})$$
CS402:
$$exactly(1,\{cr_{CS402,r_1},cr_{CS402,r_2},cr_{CS402,lab1},cr_{CS402,lab2}\})$$
$$exactly(1,\{cr_{CS402,lab1},cr_{CS402,lab2}\})$$
CS408:
$$exactly(1,\{cr_{CS408,r_1},cr_{CS408,r_2},cr_{CS408,lab1},cr_{CS408,lab2}\})$$
$$exactly(1,\{cr_{CS408,lab1},cr_{CS408,lab2}\})$$

\item Number of lectures. The clauses are grouped by courses:\\
CS101:

$exactly(2,\{ct_{CS101,t_1},ct_{CS101,t_2},ct_{CS101,t_3},ct_{CS101,t_4},\\ct_{CS101,t_5}\})$\\

CS202:

$exactly(2,\{ct_{CS202,t_1},ct_{CS202,t_2},ct_{CS202,t_3},ct_{CS202,t_4},\\ct_{CS202,t_5}\})$\\

M271:

$exactly(2,\{ct_{M271,t_1},ct_{M271,t_2},ct_{M271,t_3},ct_{M271,t_4},\\ct_{M271,t_5}\})$\\

CS304:

$exactly(2,\{ct_{CS304,t_1},ct_{CS304,t_2},ct_{CS304,t_3},ct_{CS304,t_4},\\ct_{CS304,t_5}\})$\\

CS305:

$exactly(2,\{ct_{CS305,t_1},ct_{CS305,t_2},ct_{CS305,t_3},ct_{CS305,t_4},\\ct_{CS305,t_5}\})$\\

CS402:

$exactly(2,\{ct_{CS402,t_1},ct_{CS402,t_2},ct_{CS402,t_3},ct_{CS402,t_4},\\ct_{CS402,t_5}\})$\\

CS408:

$exactly(2,\{ct_{CS408,t_1},ct_{CS408,t_2},ct_{CS408,t_3},ct_{CS408,t_4},\\ct_{CS408,t_5}\})$
\end{enumerate}
\subsubsection{Solution}
The following table can be created by feeding the encoding clauses to a Max-SAT solver and looking at the truth assignments of the $ct$ and the $cr$ variables in the output model. 
\begin{center}
\tiny{
\begin{tabular}{|c|c|c|c|c|c|}
\hline \textbf{Rooms$\slash$Timeslots} & $t_1$ & $t_2$ & $t_3$ & $t_4$ & $t_5$ \\ 
\hline $r_1$ &  &  & CS305 lect. & CS202 lect. &  \\ 
\hline $r_2$ & M271 lect. & M271 sec. & CS101 lect. & CS304 lect. & CS408 lect. \\ 
\hline lab1 &  & CS101 lect. & CS202 lab &  & CS305 lab \\ 
\hline lab2 & CS402 lect. & CS304 lab & CS402 lab & CS408 lab & CS101 lect. \\ 
\hline 
\end{tabular}
}
\end{center}
\section{Conclusion and future work}
This paper showed how to encode the timetabling problem of the department of mathematics, Cairo university using Max-SAT technology. The constraints specific to the problem have been encoded in CNF and all the clauses encoding each constraint can be grouped together to form one CNF formula to be fed to a Max-SAT solver and generate a solution. The timetable can be created by checking the output model for the truth assignments of the basic variables of the encoding. A detailed example has been given to illustrate the workings of our method.

The encoding of some constraints can be further enhanced such as room capacity. It is sometimes possible to split the students of a section or a laboratory into two groups if the number of students exceeds the room capacity. Each group can have its own group of teaching assistants. This idea is not very applicable for lectures as the number of teachers is small compared to that of teaching assistants.

We are also interested in the problem of determining the minimum amount of a certain resource (e.g., rooms, timeslots) necessary to have a conflict-free timetable.



%
\bibliographystyle{plain}
\bibliography{references}

\end{document}